\documentclass[letterpaper, 10 pt, conference]{ieeeconf}  % Comment this line out if you need a4paper

\IEEEoverridecommandlockouts                              % This command is only needed if 
\usepackage{cite}
\usepackage{amsmath}
\usepackage{multirow}
\usepackage{amssymb}
\usepackage{graphicx}
\usepackage{booktabs}
\makeatletter
\let\NAT@parse\undefined
\makeatother
\usepackage[colorlinks,
            linkcolor=black,
            anchorcolor=black,
            citecolor=black
            ]{hyperref}
\title{\LARGE \bf
MonoDiff9D: Monocular Category-Level 9D Object\\Pose Estimation via Diffusion Model
}

\author{Jian Liu, Wei Sun, Hui Yang, Jin Zheng, Zichen Geng, Hossein Rahmani, and Ajmal Mian
\thanks{Jian Liu, Wei Sun, and Hui Yang are with the National Engineering Research Center for Robot Visual Perception and Control Technology, College of Electrical and Information Engineering, and also with the School of Robotics, Hunan University, Changsha 410082, China. (e-mail: \{jianliu, wei\_sun, huiyang\}@hnu.edu.cn)}
\thanks{Jin Zheng is with the Central South University, Changsha, 410082, China. (e-mail: zheng.jin@csu.edu.cn)}
\thanks{Hossein Rahmani is with the Lancaster University, LA1 4YW, United Kingdom. (e-mail: h.rahmani@lancaster.ac.uk)}
\thanks{Zichen Geng and Ajmal Mian are with The University of Western Australia, WA 6009, Australia. (e-mail: zen.geng@research.uwa.edu.au, ajmal.mian@uwa.edu.au)}
}

\begin{document}

\maketitle
\thispagestyle{empty}
\pagestyle{empty}

%%%%%%%%%%%%%%%%%%%%%%%%%%%%%%%%%%%%%%%%%%%%%%%%%%%%%%%%%%%%%%%%%%%%%%%%%%%%%%%%
\begin{abstract}

Object pose estimation is a core means for robots to understand and interact with their environment. For this task, monocular category-level methods are attractive as they require only a single RGB camera. However, current methods rely on shape priors or CAD models of the intra-class known objects. We propose a diffusion-based monocular category-level 9D object pose generation method, MonoDiff9D.
Our motivation is to leverage the probabilistic nature of diffusion models to alleviate the need for shape priors, CAD models, or depth sensors for intra-class unknown object pose estimation.
We first estimate coarse depth via DINOv2 from the monocular image in a zero-shot manner and convert it into a point cloud. We then fuse the global features of the point cloud with the input image and use the fused features along with the encoded time step to condition MonoDiff9D. Finally, we design a transformer-based denoiser to recover the object pose from Gaussian noise. Extensive experiments on two popular benchmark datasets show that MonoDiff9D achieves state-of-the-art monocular category-level 9D object pose estimation accuracy without the need for shape priors or CAD models at any stage. Our code will be made public at \href{https://github.com/CNJianLiu/MonoDiff9D}{https://github.com/CNJianLiu/MonoDiff9D}.

\end{abstract}

%%%%%%%%%%%%%%%%%%%%%%%%%%%%%%%%%%%%%%%%%%%%%%%%%%%%%%%%%%%%%%%%%%%%%%%%%%%%%%%%
\section{INTRODUCTION}\label{sec:intro}
Nine-degrees-of-freedom (9D) object pose estimation aims to predict the 3D translation, 3D rotation, and 3D size of an object relative to the camera \cite{Gen6D, STG6D, GigaPose, liu2023fine, cai2024open, liu2024survey}. It plays a crucial role in robotic 3D scene understanding \cite{manhardt2019roi, fu20226d, remus2023i2c, NOPE, UniMODE, DGPF6D}. Early object pose estimation methods are mainly instance-level methods \cite{PVNet, PVN3D, DenseFusion, FFB6D, GDR-Net, HFF6D}. They achieve good results on specific object instances, but have poor generalization ability. Recently, category-level methods have received extensive research attention due to their ability to generalize to intra-class unknown objects without requiring CAD models of the objects \cite{CAPTRA, BundleTrack, MH6D, CR-Net, PhoCaL, HouseCat6D, AG-Pose}.

\par As one of the pioneers in category-level research, Wang \emph{et al.} \cite{NOCS} proposed a Normalized Object Coordinate Space (NOCS) for the standardized representation of a category of objects, and used the Umeyama algorithm \cite{Umeyama} to match the predicted NOCS shape and the observed object point cloud to solve the pose. Following this, several RGBD-based methods were proposed that fuse the 2D texture and 3D geometric features of the observed object for pose estimation \cite{SPD, SGPA, SSP-Pose, DualPoseNet, IST-Net, VI-Net}. In addition, some RGBD-based domain adaptation and domain generalization methods have been introduced \cite{SSC6D, UDA-COPE, Wild6D, TTA-COPE, DPDN}, which aim to reduce the reliance on real-world annotated data. Besides these RGBD-based methods, some depth-based methods have been proposed \cite{FS-Net, GPV-Pose, HS-Pose, SAR-Net}, which further focus on the geometric features of the observed object and then directly regress the object pose. Although these RGBD and depth-based methods achieve superior performance, they exhibit excessive dependence on depth information. Since depth sensors are energy-consuming for robots and most mobile devices (such as mobile phones, tablets, laptops, \emph{etc}.) are not equipped with them, achieving monocular category-level pose estimation holds significant importance across diverse applications.

\begin{figure}[t!]
\centering
\includegraphics[width=\linewidth]{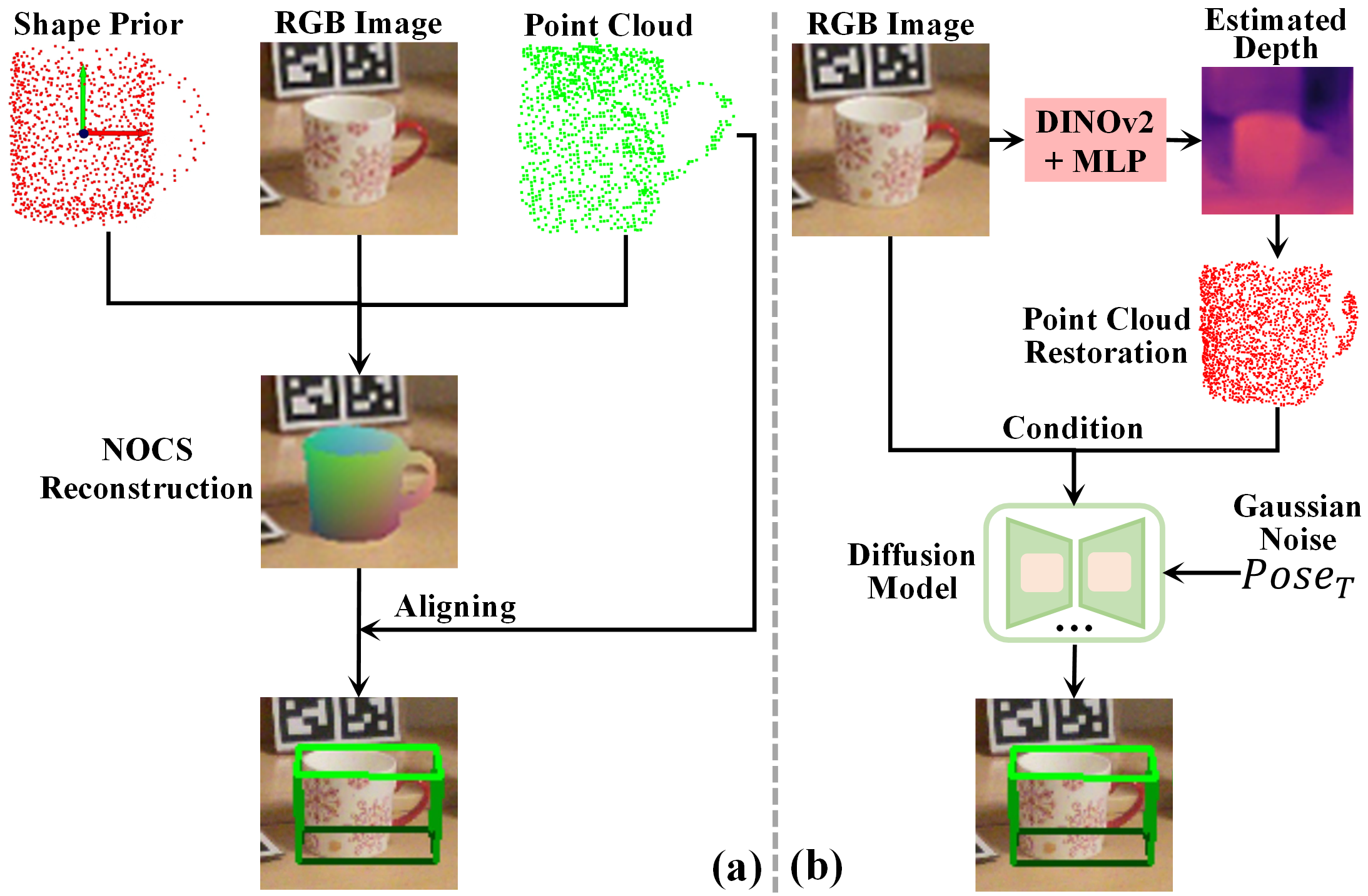}
\vspace{-1.5em}
\caption{Comparison between (a) shape prior-based RGBD methods and (b) the proposed MonoDiff9D. (a) leverages shape prior, RGB image, and point cloud to reconstruct the NOCS representation of the object, and uses NOCS shape alignment to solve the object pose. (b) first estimates the coarse object point cloud based on DINOv2 \cite{DINOv2}, and then uses the RGB image and the point cloud as conditions to guide the diffusion model for recovering the object pose.}
\label{Fig1}
\vspace{-1.5em}
\end{figure}

\par To avoid the need for depth sensor, some monocular RGB-based methods have been proposed \cite{Synthesis, MSOS, OLD-Net, DMSR, manhardt2020cps++}. Synthesis \cite{Synthesis} and MSOS \cite{MSOS} estimate object pose from a single RGB image, however, their performance is greatly limited since a monocular RGB image cannot accurately represent the 3D location and geometric information. More recently, OLD-Net \cite{OLD-Net} and DMSR \cite{DMSR} first leverage a single RGB image to estimate the depth or sketch of the observed object, and then combine the estimated geometric information and shape priors to reconstruct the NOCS shape of the intra-class unknown object. Next, the 9D object pose is solved through the Umeyama \cite{Umeyama} or RANSAC+Perspective-n-Point (PnP) \cite{DMSR} algorithm. Although these methods improve accuracy, they either require the Umeyama algorithm to solve the pose, or the RANSAC algorithm and the PnP procedure to remove outliers and solve the pose, respectively, making them non-differentiable. Furthermore, they all rely on using shape priors or the CAD models of intra-class known objects during training, which need extensive manual effort to obtain.

\par To address the aforementioned challenges, inspired by the prior work Diff9D \cite{Diff9D}, we propose a differentiable and shape prior-free monocular category-level 9D object pose estimation method, coined MonoDiff9D. Our aim is to exploit the probabilistic nature of the diffusion model to perform intra-class unknown object pose estimation, even in the absence of shape priors, CAD models, or depth sensors.
Since the depth image/point cloud can represent the location and geometric information of the object, it is the most critical input for object pose estimation. In particular, estimating 3D object translation from a single RGB image is tricky. Based on this, we propose to leverage the Large Vision Model (LVM) DINOv2 \cite{DINOv2} to perform coarse depth estimation from the input RGB image in a zero-shot manner and convert it into a point cloud, thereby introducing the location and geometric information of the observed object to condition the learning of the diffusion model. Since there is still a gap between the DINOv2 estimated and the actual object point clouds, and the complete point cloud may not be restored, we then leverage the self-attention and cross-attention mechanisms to fuse the global features of the RGB image and the estimated point cloud, so that it can pay more attention to the pose-sensitive features of the coarse point cloud. Finally, we design a transformer-based denoiser following the Denoising Diffusion Implicit Model (DDIM) \cite{DDIM} scheduler to recover the object pose from Gaussian noise. A comparison between the shape prior-based RGBD methods and the proposed MonoDiff9D is shown in Fig. \ref{Fig1}. Overall, we make the following contributions:

\begin{itemize}
\item We propose a diffusion model-based monocular category-level 9D object pose generation framework, designed to enhance generalization to intra-class unknown objects when shape priors and CAD models of intra-class known objects are unavailable.

\item We propose LVM-based conditional encoding to introduce coarse location and geometric prompts to condition the learning of the single RGB-based pose diffusion process, and demonstrate that LVM can effectively adapt to this framework without any fine-tuning.

\item We demonstrate on two widely used benchmarks that object pose can be recovered from Gaussian noise with state-of-the-art (SOTA) accuracy in near real-time. %Moreover, MonoDiff9D has stronger generalization in the wild than regression-based methods.
\end{itemize}

\begin{figure*}[t!]
\centering
\includegraphics[width=\textwidth]{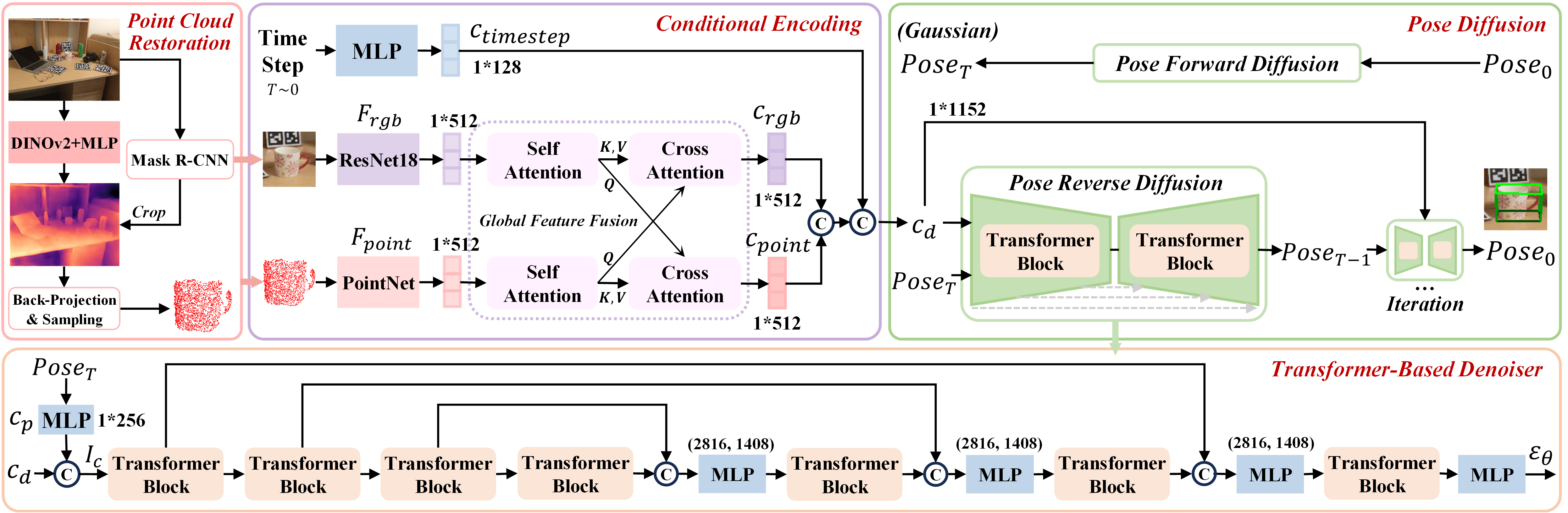}
\vspace{-1.5em}
\caption{Overall workflow of the proposed MonoDiff9D, which includes three parts: point cloud restoration, conditional encoding, and pose diffusion. 
%(1) We estimate the coarse depth image based on DINOv2 \cite{DINOv2}, segment the image using Mask R-CNN \cite{MaskR-CNN}, use the segmented mask to crop the depth image, and convert the cropped depth image into a point cloud. (2) We use MLP to encode the time step, and leverage self-attention and cross-attention mechanisms to fuse the global features of the observed RGB image and the point cloud. These three features are concatenated to form the conditional encoding. (3) The diffusion model is used to denoise the Gaussian noise based on the encoded condition to recover the actual object pose.
}
\label{Fig2}
\vspace{-0.5em}
\end{figure*}

\section{RELATED WORK}
We review related category-level object pose estimation works by categorizing them into RGBD-based, depth-based, and RGB-based methods.
\subsection{RGBD-Based and Depth-Based Methods}
SPD \cite{SPD} first uses the category-level CAD model library to extract shape priors offline. Then, it utilizes shape prior deformation to reconstruct the NOCS shape of intra-class unknown object and leverages the Umeyama algorithm \cite{Umeyama} to solve pose. Due to the superior performance achieved by SPD, many other prior-based methods \cite{SGPA, centersnap, SSP-Pose, fsd, diffusionnocs} are subsequently proposed.
%Specifically, SGPA \cite{SGPA} adjusts the shape prior dynamically by using the structure similarity between the observed object and the shape prior. RBP-Pose \cite{RBP-Pose} aims to solve the insufficient pose-sensitive feature extraction through a geometry-guided residual object bounding box projection network. SSP-Pose \cite{SSP-Pose} proposes an end-to-end pose estimation network, avoiding the utilization of the non-differentiable Umeyama algorithm \cite{Umeyama}.
Although these prior-based methods can deal with diverse shape variations between intra-class objects, acquiring shape priors requires manual effort to build the category-level CAD model library. Hence, some prior-free methods like DualPoseNet \cite{DualPoseNet}, IST-Net \cite{IST-Net}, and VI-Net \cite{VI-Net} are introduced.
%DualPoseNet \cite{DualPoseNet} designs a dual pose encoder with refined learning of pose consistency and predicts object pose via two parallel pose decoders. IST-Net \cite{IST-Net} explores the necessity of shape priors and introduces a prior-free pipeline based on implicit space transformation. VI-Net \cite{VI-Net} decouples rotation into in-plane and viewpoint rotations to address the challenge of poor rotation estimation. SecondPose \cite{SecondPose} introduces DINOv2 to provide SE(3)-consistent semantic features for pose estimation.

\par Besides the above RGBD-based methods, there are also some depth-based methods\cite{FS-Net, GPV-Pose, HS-Pose, CPPF, SAR-Net}. FS-Net \cite{FS-Net} introduces a 3D graph convolution-based framework and performs decoupled regression for 6D pose and 3D size. %GPV-Pose \cite{GPV-Pose} enhances the learning of pose-sensitive features via geometry-guided point-wise voting. HS-Pose \cite{HS-Pose} addresses the limitations related to the translation and size invariant properties of 3D graph convolution by designing a hybrid scope feature extraction network.
GPV-Pose \cite{GPV-Pose} and HS-Pose \cite{HS-Pose} improve the model accuracy by enhancing the learning of pose-sensitive features and designing a hybrid scope feature extraction network, respectively.
%CPPF \cite{CPPF} proposes a category-level point pair feature voting mechanism to perform pose estimation in the wild. SAR-Net \cite{SAR-Net} explores the shape alignment of the observed object point cloud against its corresponding shape prior.
GenPose \cite{GenPose} introduces a score-based diffusion method for generative 6D object pose estimation. Although these depth-based methods achieve superior performance, they heavily rely on the depth information of the observed object. Due to the energy-consuming nature of depth sensors and most mobile devices are not equipped with them, achieving category-level object pose estimation through a single RGB image is significant for widespread applications.

\subsection{RGB-Based Methods}
Synthesis \cite{Synthesis} incorporates a neural synthesis module into a model fitting framework based on optimization to predict object shape and pose concurrently. MSOS \cite{MSOS} predicts the metric scale shape and the NOCS shape of the observed object, and performs a similarity transformation between them to recover the pose. However, their accuracy is greatly limited by the lack of guidance from geometric information. Lin \emph{et al.} \cite{lin2022single} introduced a keypoint-based, single-stage pipeline that operates on a single RGB image. More recently, OLD-Net \cite{OLD-Net} reconstructs object-level depth and NOCS shape directly from monocular RGB image by deforming category-level shape prior and then aligning them to estimate the object pose. DMSR \cite{DMSR} first predicts the 2.5D sketch and decouples scale recovery based on the shape prior, then reconstructs the NOCS shape for pose estimation. These methods enhance accuracy but typically employ non-differentiable techniques such as the Umeyama algorithm \cite{Umeyama} for pose recovery or a combination of RANSAC for outlier removal and the PnP approach for solving pose \cite{DMSR}. Moreover, they often rely on shape priors and CAD models of intra-class known objects during the training process.

\par Different from the above methods, our MonoDiff9D aims to leverage the LVM DINOv2 \cite{DINOv2} to provide geometric guidance to condition the Denoising Diffusion Probabilistic Model (DDPM) \cite{DDPM}, achieving differentiable and shape prior-free monocular category-level 9D object pose estimation without using any CAD models during training.

\section{WORKFLOW of MONODIFF9D} %why not 
%\subsection{Overview of MonoDiff9D}
Figure \ref{Fig2} shows an overview of our proposed model.  The goal of MonoDiff9D is to predict the 6D pose (composed of 3D translation and 3D rotation) and 3D size (axes lengths of the oriented bounding box) of the intra-class unknown object using only a single RGB image. Given an input RGB image, we first leverage the pre-trained LVM DINOv2 \cite{DINOv2} and Mask R-CNN \cite{MaskR-CNN} to generate the coarse depth image and segmented RGB image, respectively. The segmented mask is used to crop the coarse depth image. Then, we convert the cropped depth image into a point cloud via back-projection and sampling. Subsequently, we encode all input conditions (including the diffusion time step ${c_{timestep}}$, RGB image ${c_{rgb}}$, and restored point cloud ${c_{point}}$ as shown in Fig. \ref{Fig2}) to provide conditional guidance for accurate reverse diffusion. Specifically, we use MLP to encode the diffusion time step, and leverage the lightweight ResNet18 \cite{ResNet} and PointNet \cite{PointNet} to extract the global features of the observed RGB image and the restored point cloud, respectively. We employ self-attention and cross-attention mechanisms to fuse these two global features. Next, the encoded ${c_{timestep}}$, ${c_{rgb}}$, and ${c_{point}}$ are concatenated to form the conditional encoding ${c_{d}}$ of the diffusion model. Finally, the DDIM \cite{DDIM} is used to denoise the Gaussian noise and recover the actual object pose through a transformer-based denoiser.

\subsection{Point Cloud Restoration}
Point cloud is the most critical input for 9D object pose estimation since it represents the location and geometric information of the observed object. In particular, estimating the 3D object translation from a monocular RGB image is challenging. To introduce the location and geometric information from a single RGB image to condition the learning of the diffusion model, we propose to leverage the pre-trained DINOv2 \cite{DINOv2} to perform coarse depth estimation from the RGB image in a zero-shot manner. We elaborate on our choice of using  DINOv2 instead of other depth estimators \cite{Depthanything, Zoedepth} in Sec. \ref{Depth Estimators}.

\par The workflow of the point cloud restoration is shown in Fig. \ref{Fig2}. Specifically, we first employ DINOv2 to estimate the coarse depth image corresponding to the observed RGB image. Then, we utilize the Mask R-CNN \cite{MaskR-CNN} to perform instance segmentation on the RGB image. Using the segmented mask, we crop the coarse depth image accordingly. \textcolor{black}{Subsequently, the cropped depth image is transformed into a point cloud and downsampled to 1024 points \cite{SGPA, DPDN}.}

\subsection{Conditional Encoding}\label{Conditional Encoding}
\par Adding conditional encoding introduces prompts that make the learning of the diffusion model more accurate \cite{GenPose, DiffPose, LatentDiffusion}. Based on this, taking the time step $T$ as an example, we first use MLP to encode the time step $T$. Since the RGB image and point cloud can provide texture and geometric information of the observed object, respectively, these are important inputs for 9D object pose estimation. Conditional encoding of the RGB image and point cloud can effectively make diffusion model learning more robust.

\par The overall structure of the conditional encoding is shown in Fig. \ref{Fig2}. We utilize the lightweight ResNet \cite{ResNet} with the last fully connected layer removed and the PointNet \cite{PointNet} to extract the global features of RGB image and point cloud, respectively. Different from the standard PointNet \cite{PointNet}, we only leverage it to extract the global feature as the condition for the efficiency of the diffusion process. Given the disparity between the object point cloud reconstructed using DINOv2 and the actual object point cloud, we utilize self-attention and cross-attention mechanisms to fuse the RGB global feature ${F_{rgb}}$ and the point cloud global feature ${F_{point}}$. This enables our model to pay more attention to the pose-sensitive features of the restored coarse point cloud. The network structures of the self-attention and cross-attention mechanisms are composed of a transformer block, similar to ViT \cite{ViT}. Their most critical part is the multi-head attention. Taking ${F_{rgb}}$ as an example, it can be expressed as:
\begin{equation}\label{equation1}
Attention\left( {F_{rgb}} \right) = F{C_{Cat}}\left( {Softmax\left( {{Q}{K{^ \top}}/\sqrt d } \right){V}} \right),
\end{equation}
where $Q$, $K$, and $V$ represent query, key, and value, respectively. For self-attention, $Q$, $K$, and $V$ all come from $F_{rgb}$. For cross-attention, $Q$ is from $F_{point}$, and $K$ and $V$ are from $F_{rgb}$. $\top$ denotes the matrix transpose operation. $d$ represents the feature dimension of each attention head. $F{C_{Cat}}\left( {} \right)$ denotes first concatenating the feature of all heads, and then feeding the concatenated features to a fully connected layer.
Finally, the encoded time step ${c_{timestep}}$, RGB image ${c_{rgb}}$, and point cloud ${c_{point}}$ are concatenated to form the conditional encoding ${c_{d}}$ of the diffusion model as follows:
\begin{equation}\label{equation2}
{c_d} = Cat{\left( {{c_{timestep}},{c_{rgb}},{c_{point}}} \right)}.
\end{equation}

\subsection{Pose Diffusion}
We leverage the diffusion model because the depth image estimated by DINOv2 \cite{DINOv2} may have an error of $10 \sim 20cm$, and the complete object may not be recovered. When the depth sensors are unavailable, regression-based models such as OLD-Net \cite{OLD-Net} and DMSR \cite{DMSR} require shape priors and the CAD models of intra-class known objects as additional supervision signals to provide more refined geometric guidance during training. In contrast, diffusion models have strong robustness to recover high-quality outputs from uncertain and noisy conditions \cite{LatentDiffusion}. Hence, we do not need to use any shape priors and CAD object models.

\par The pose diffusion consists of two stages, each characterized as a Markov chain. Specifically, the first stage is forward diffusion, where Gaussian noise with predefined mean and variance is gradually added to the ground-truth object pose. Subsequently, the reverse diffusion stage employs a neural network to progressively denoise Gaussian noise, thus reconstructing the actual object pose. 

\subsubsection{Forward Diffusion}
Given a ground-truth object pose ${p_0}$, which conforms to the pose distribution ${{p}_0} \sim q\left( p \right)$, we define a forward diffusion process where Gaussian noise ${\cal N}\left( {0 ,\textbf{\emph{I}}} \right)$ is gradually added to the ground-truth object pose in $T$ time steps, producing a sequence of noisy poses ${p_1}, \cdot  \cdot  \cdot ,{p_T}$. The sequence is controlled by a cosine-schedule $\left\{ {{\beta _t} \in \left( {0,1} \right)} \right\}_{t = 1}^T$ and ${\beta _1} < {\beta _2} <  \cdot  \cdot  \cdot  < {\beta _T}$ \cite{ImprovedDDPM}. Following DDPM \cite{DDPM}, the step-by-step diffusion process follows the Markov chain assumption:
\begin{equation}\label{equation4}
q\left( {{p_{1:T}}|{p_0}} \right) = \prod\limits_{t = 1}^T {q\left( {{p_t}|{p_{t - 1}}} \right)}.
\end{equation}
Specifically, ${p_t}$ can be defined as:
\begin{equation}\label{equation5}
{p_t} = \sqrt {{\beta _t}} {\varepsilon _t} + \sqrt {1 - {\beta _t}} {p_{t - 1}},
\end{equation}
where ${\varepsilon _t}$ is a randomly sampled standard Gaussian noise at time step $t$. Let ${\alpha _t} = 1 - {\beta _t}$ and ${{\bar \alpha }_t} = \prod\nolimits_{i = 1}^t {{\alpha _i}}$ and iterate Eq. \ref{equation5}, the forward pose diffusion process from ${p_{0}}$ to ${p_t}$ can be expressed as:
\begin{equation}\label{equation6}
q\left( {{p_t}|{p_0}} \right) = {\cal N}\left( {{p_t};\sqrt {{{\bar \alpha }_t}} {p_0},\left( {1 - {{\bar \alpha }_t}} \right)\textbf{\emph{I}}} \right).
\end{equation}

\subsubsection{Reverse Diffusion}
As shown in Fig. \ref{Fig2}, the reverse diffusion process aims to recover the actual object pose from a standard Gaussian noise input ${p_T} \sim {\cal N}\left( {0,I} \right)$ conditioned on ${c_{d}}$. However, it is difficult to obtain $q\left( {{p_{t - 1}}|{p_t}} \right)$ directly, so we run the reverse diffusion process by learning a denoising model ${r_\theta }$ to approximate this conditional probability as:
\begin{equation}\label{equation7}
{r_\theta }\left( {{p_{t - 1}}|{p_t}} \right) = {\cal N}\left( {{p_{t - 1}};{\mu _\theta }\left( {{p_t},t,{c_{d}}} \right),\sum\nolimits_\theta  {\left( {{p_t},t} \right)} } \right),
\end{equation}
where ${c_{d}}$ represents the condition (see Sec. \ref{Conditional Encoding} for details). Following DDPM \cite{DDPM} to use Bayes' theorem transform Eq. \ref{equation7}, then we can get the variance $\sum\nolimits_\theta  {\left( {{p_t},t} \right)} = \frac{{1 - {{\bar \alpha }_{t - 1}}}}{{1 - {{\bar \alpha }_t}}} \cdot {\beta _t} \cdot I$ and the mean
\begin{equation}\label{equation8}
{\mu _\theta }\left( {{p_t},t,{c_{d}}} \right) = \frac{1}{{\sqrt {{\alpha _t}} }}\left( {{p_t} - \frac{{1 - {\alpha _t}}}{{\sqrt {1 - {{\bar \alpha }_t}} }}{\varepsilon _\theta }\left( {{p_t},t,{c_{d}}} \right)} \right).
\end{equation}
Specifically, we first use MLP to encode the input $p_t$ of the diffusion model to obtain the pose feature $c_p$, and then concatenate it with $c_d$ to form the input of the denoiser as follows:
\begin{equation}\label{equation8_1}
I_{c} = Cat{\left( {c_{d},{c_{p}}} \right)^ \top}.
\end{equation}
Since $I_c$ is a one-dimensional vector, we design a transformer-based U-Net as the denoiser. By doing so, each feature channel in $c_p$ can be globally associated with $c_d$, thus contributing to more robust diffusion. The detailed network structure is shown in Fig. \ref{Fig2}. We utilize transformer blocks that are identical to those used in ViT \cite{ViT}.

\par To improve the running speed of reverse diffusion, we leverage the DDIM scheduler \cite{DDIM} to reduce the number of sampling time steps.
%, reducing the number of sampling time steps from $T$ to $S$ via taking a sample every $T/S$ time steps. Now, the new sampling schedule is $\left\{ {{\tau _1}, \cdot  \cdot  \cdot ,{\tau _S}} \right\}$.
Overall, the reverse diffusion aims to estimate the pose noise ${\varepsilon _\theta }$ by learning a neural network conditioned on ${c_{d}}$. Then, we can leverage the DDIM \cite{DDIM} scheduler to denoise, recovering the actual object pose.

\subsection{Training Protocol}
Since ${{\varepsilon _t}}$ can be easily obtained at training time, we use it as the ground truth to supervise the learning of the denoising model. Specifically, the denoising loss term can be parameterized to minimize the difference between ${{\varepsilon _\theta }\left( {{p_t},t,{c_{d}}} \right)}$ and ${{\varepsilon _t}}$ as follows:
\begin{align}
\begin{split}
\label{equation10}
%\begin{array}{l}
%\begin{array}{l}
{L}& = {\mathbb{E}_{t \sim \left[ {1,T} \right],{p_0},{\varepsilon _t}}}\left[ {{{\left\| {{\varepsilon _t} - {\varepsilon _\theta }\left( {{p_t},t,{c_{d}}} \right)} \right\|}^2}} \right].
 \end{split}
%\end{array}
%\end{array}
\end{align}

\par For the representation of the 9D object pose, we follow \cite{IST-Net, DPDN, GenPose} to directly use the translation, rotation, and size matrices $t$, $R$, and $s$ to represent the translation, rotation, and size, respectively.

\section{EXPERIMENTS}
\subsection{Evaluation Datasets and Metrics}\label{Datasets}
\noindent\textbf{Datasets:}
We follow previous monocular category-level object pose estimation methods \cite{Synthesis, MSOS, OLD-Net, DMSR} to use the CAMERA25 and REAL275 datasets \cite{NOCS} for evaluation. CAMERA25 is a large synthetic dataset including 1085 instances from 6 categories of objects. REAL275 is a real-world dataset containing 42 instances from the same categories of objects as in CAMERA25. The data split and training mode are identical to the compared methods \cite{Synthesis, MSOS, OLD-Net, DMSR}.

\begin{table*}[t!]
\renewcommand\arraystretch{1.1}
\newcommand{\tabincell}[2]{\begin{tabular}{@{}#1@{}}#2\end{tabular}}
  \centering
  \caption{Comparison on the CAMERA25 and REAL275 datasets in mean average precision (mAP) metrics of $Io{U_{3D}}$ ($\%$), $n^\circ$ ($\%$), and $m{\rm{cm}}$ ($\%$). "\checkmark" and "-" denote required and not required. The best results are bolded.}
  \vspace{-0.7em}
  \setlength{\tabcolsep}{4.6pt}{
    \begin{tabular}{c|cc|ccccc|ccccc}
    \hline
    \multirow{2}[2]{*}{Method} & \multirow{2}[2]{*}{Shape Prior} & \multirow{2}[2]{*}{CAD Model} & \multicolumn{5}{c}{CAMERA25} & \multicolumn{5}{|c}{REAL275}\\
\cline{4-13}   &  &  & ${\rm{3}}{{\rm{D}}_{50}}$ & ${\rm{3}}{{\rm{D}}_{75}}$ & \tabincell{c}{$10{\rm{cm}}$} & \tabincell{c}{$10^\circ$} & \tabincell{c}{$10^\circ$$10{\rm{cm}}$} & ${\rm{3}}{{\rm{D}}_{50}}$ & ${\rm{3}}{{\rm{D}}_{75}}$ & \tabincell{c}{$10{\rm{cm}}$} & \tabincell{c}{$10^\circ$} & \tabincell{c}{$10^\circ$$10{\rm{cm}}$}\\
    \hline
    Synthesis \cite{Synthesis}  & - & \checkmark & - & - & - & - & - & -  & -  & 34.0  & 14.2  & 4.8 \\
    MSOS \cite{MSOS} & -  & \checkmark & 32.4  & 5.1  & 29.7 & 60.8 & 19.2 & 23.4  & 3.0  & 39.5  & 29.2  & 9.6\\
    OLD-Net \cite{OLD-Net} & \checkmark  & \checkmark & 32.1  & 5.4   & 30.1 & 74.0   & 23.4 & 25.4  & 1.9   & 38.9 & 37.0   & 9.8\\
    DMSR \cite{DMSR} & \checkmark & \checkmark  & 34.6  & 6.5   & 32.3 & \textbf{81.4}   & 27.4 & 28.3  & 6.1   & 37.3 & \textbf{59.5}   & 23.6\\
    MonoDiff9D & - & - & \textbf{35.2}  & \textbf{6.7} & \textbf{33.6} & 80.1 & \textbf{28.2} & \textbf{31.5}  & \textbf{6.3} & \textbf{41.0} & 56.3 & \textbf{25.7}\\
    \bottomrule
    \end{tabular}}
  \label{table1}
  \vspace{-1.5em}
\end{table*}

\begin{figure}[t!]
\centering
\includegraphics[width=\linewidth]{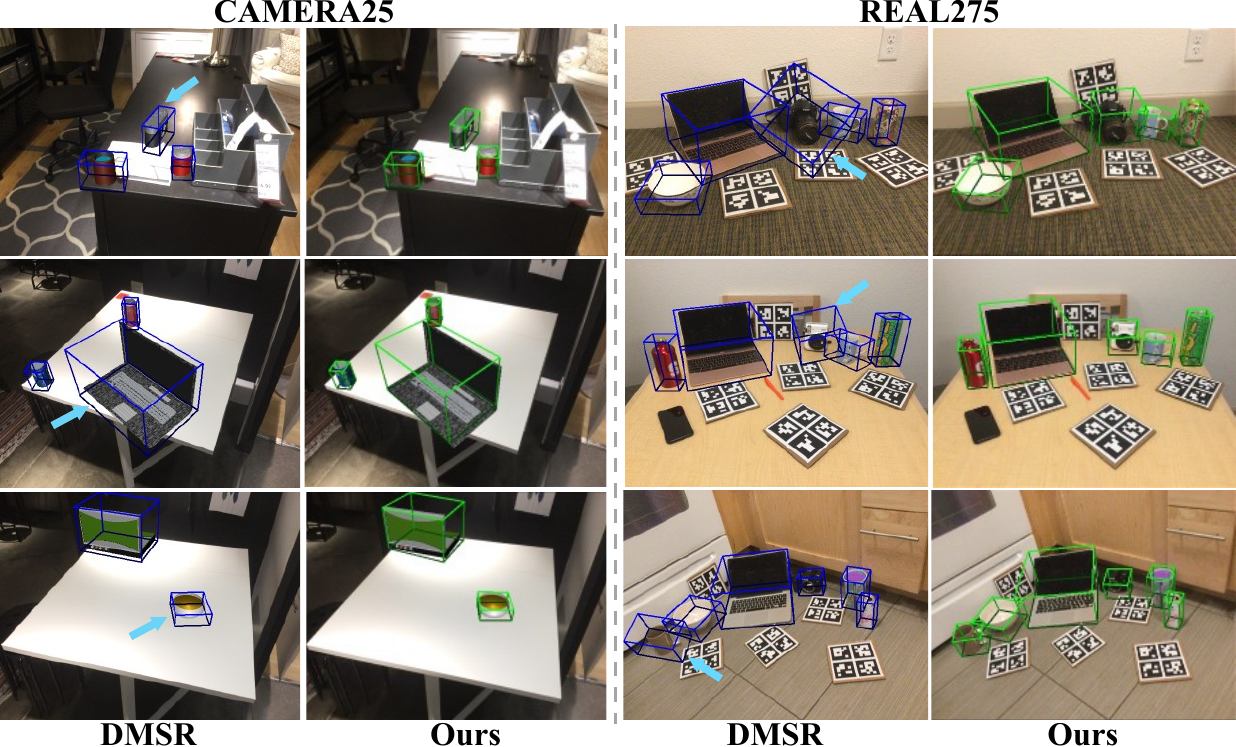}
\vspace{-2em}
\caption{Qualitative results comparison with SOTA method DMSR \cite{DMSR} on the CAMERA25 (left) and REAL275 (right) datasets. The blue arrows point to areas where DMSR has significantly higher error than our method.}
\label{Fig4}
\vspace{-1.5em}
\end{figure}

\noindent\textbf{Metrics:}
Following the comparison methods \cite{Synthesis, MSOS, OLD-Net, DMSR}, we choose the widely used $n^\circ m{\rm{cm}}$ and 3D Intersection-over-Union ($Io{U_{3D}}$) metrics for evaluation. $n^\circ m{\rm{cm}}$ directly denotes the predicted rotation and translation errors. $Io{U_{3D}}$ represents the percentage of intersection and union of the ground-truth and the estimated 3D bounding boxes, which is mainly used to evaluate the size estimation. We follow DMSR \cite{DMSR} to select 50$\%$ and 75$\%$ as the thresholds of $Io{U_{3D}}$  and choose $10^\circ$, $10{\rm{cm}}$, and $10^\circ 10{\rm{cm}}$ to evaluate the translation and rotation estimation.

\subsection{Implementation Details}
%In the reverse diffusion process, we set the number of sampling time steps $S$ to 2 (see Table \ref{table5} for reasoning) by leveraging the DDIM \cite{DDIM} scheduler to improve the running speed. The resolution of the initial RGBD image is $640 \times 480$. We first perform instance segmentation for the initial image via Mask R-CNN \cite{MaskR-CNN}, then use the segmented mask to crop the depth image as well and scale the RGB image to $192 \times 192$ to reduce further computations.
Since we set the number of attention heads of all the transformer blocks in Fig. \ref{Fig2} to 16 through experiments, the feature dimension $d$ in Eq. \ref{equation1} is set to 32 (calculated by ${{512} \mathord{\left/ {\vphantom {{512} 16}} \right. \kern-\nulldelimiterspace} 16}$). The diffusion time step $T$ in Eq. \ref{equation4} is set to 1000. The learning rate is cyclically adjusted between $1{e^{ - 6}}$ and $1{e^{ - 4}}$ through the CyclicLR function \cite{CyclicLR}, and the step size of each cycle is set to 20K. We train for 30 cycles, and the batch size of each step is set to 48. Experiments are conducted using a single NVIDIA 3090 GPU. The pose diffusion process achieves a speed of \emph{26.3 frames per second}.

\subsection{Comparisons with State-of-the-Art Methods}
We follow previous SOTA methods \cite{MSOS, OLD-Net, DMSR} to use the CAMERA25 and REAL275 datasets for evaluation. To train and test MonoDiff9D on these two datasets, we extend them with DINOv2 \cite{DINOv2} (approximately 276K images) in a zero-shot manner. For a fair comparison, MonoDiff9D and all the comparison methods are provided with the same segmentation results (\emph{i.e.}, segmented by Mask R-CNN \cite{MaskR-CNN}).

\subsubsection{CAMERA25 Dataset}
We compare MonoDiff9D with three SOTA methods on the CAMERA25 dataset.
%MonoDiff9D and all the comparison methods are provided with exactly the same segmentation results (\emph{i.e.}, segmented by Mask R-CNN \cite{MaskR-CNN}) for a fair comparison.
The quantitative comparison results are shown in Table \ref{table1}. MonoDiff9D outperforms MSOS \cite{MSOS} and OLD-Net \cite{OLD-Net} by 19.3$\%$ and 6.1$\%$ on $10^\circ$ metric, respectively. Since MSOS \cite{MSOS} lacks the guidance of geometric information, it has lower rotation accuracy. In addition, MonoDiff9D outperforms OLD-Net \cite{OLD-Net} and DMSR \cite{DMSR} by 4.8$\%$ and 0.8$\%$ on $10^\circ$$10{\rm{cm}}$ metric, respectively. It is worth noting that these two comparison methods need to use shape priors and the CAD models of intra-class known objects during training, while MonoDiff9D outperforms them without using these. Some qualitative results are shown in Fig. \ref{Fig4}. We can see that MonoDiff9D has a more refined translation and size estimation performance.

\begin{table}[t!]
\renewcommand\arraystretch{1}
\newcommand{\tabincell}[2]{\begin{tabular}{@{}#1@{}}#2\end{tabular}}
  \centering
  \caption{Ablation study of depth estimators and their suitability for diffusion model on the REAL275 dataset.}
  \vspace{-0.7em}
  \setlength{\tabcolsep}{4.6pt}{
      \begin{tabular}{c | c c c c c}
    \toprule
Method & ${\rm{3}}{{\rm{D}}_{50}}$ & ${\rm{3}}{{\rm{D}}_{75}}$ & \tabincell{c}{$ 10{\rm{cm}}$} & \tabincell{c}{$10^\circ$} & \tabincell{c}{$10^\circ$$ 10{\rm{cm}}$} \\
    \midrule
    %HS-Pose*\cite{HS-Pose}  &  82.1  &  74.7  &  99.6  &  85.4   &  83.7   \\
    DINOv2 + HS-Pose \cite{HS-Pose}  &  19.0  &  1.8  &  27.6  &  42.2   &  12.1   \\
    DINOv2 + IST-Net \cite{IST-Net}  &  30.9  &  3.3  &  40.7  &  43.7   &  17.3   \\
    ZoeDepth + MonoDiff9D   &  15.6   &  1.5     &  24.3   &  \textbf{57.8}  &  14.0  \\
    DINOv2 w/o Diffusion   &  21.6   &  2.5     &  30.4   &  52.9  &  19.3  \\
    DINOv2 + MonoDiff9D   &  \textbf{31.5}   &  \textbf{6.3}     &  \textbf{41.0}   &  56.3  &  \textbf{25.7}  \\
    \bottomrule
    \end{tabular}}
  \label{table3}
  \vspace{-1.5em}
\end{table}

\subsubsection{REAL275 Dataset}
We evaluate the performance of our MonoDiff9D on the real-world REAL275 dataset. The quantitative comparison results are shown in Table \ref{table1}. MonoDiff9D outperforms Synthesis \cite{Synthesis} and MSOS \cite{MSOS} by 42.1$\%$ and 27.1$\%$ on $10^\circ$ metric, respectively. Since both Synthesis \cite{Synthesis} and MSOS \cite{MSOS} lack the guidance of geometric information, their rotation estimation accuracy in the real world is much lower than that of methods guided by geometric information. Furthermore, MonoDiff9D outperforms OLD-Net \cite{OLD-Net} and DMSR \cite{DMSR} by 6.1$\%$ and 3.2$\%$ on 50$\%$ $Io{U_{3D}}$ metric, respectively. Note that both OLD-Net \cite{OLD-Net} and DMSR \cite{DMSR} rely on using shape priors and the CAD models of intra-class known objects during training, which require extensive manual efforts to obtain. MonoDiff9D performs better without using shape priors or CAD models during training. Some qualitative results are shown in Fig. \ref{Fig4}. We can see that MonoDiff9D has better size estimation accuracy for some challenging objects (\emph{e.g.}, camera and mug).

\subsection{Ablation Studies and Discussions}
%We first conduct ablation studies on conditional encoding, pose diffusion process, and depth estimators to investigate their impact on MonoDiff9D. Then, we conduct additional studies on the generalization ability of MonoDiff9D in the wild.

\subsubsection{Depth Estimators and their Suitability for Diffusion Model}\label{Depth Estimators}
%We perform an additional ablation study to demonstrate the suitability of the diffusion model for the task at hand.
First, we take the SOTA category-level 9D object pose estimation methods HS-Pose (depth-based) \cite{HS-Pose} and IST-Net (RGBD-based) \cite{IST-Net} and replace their depth data with that restored by DINOv2 \cite{DINOv2} during training as well as inference while keeping all other experimental settings unchanged. Detailed results are shown in Table \ref{table3}. We can see that both HS-Pose and IST-Net do not perform well since the depth images from DINOv2 contain an error of $10 \sim 20cm$ and the complete object is not always restored. Our method
%outperforms the regression-based HS-Pose \cite{HS-Pose} and IST-Net \cite{IST-Net} because it is based on diffusion, which
is more robust to uncertain and noisy conditions. Additionally, other depth estimators have been proposed recently, such as Depth Anything \cite{Depthanything} and ZoeDepth \cite{Zoedepth}. Depth Anything mainly estimates the relative depth and does not perform metric depth estimation which is required by our model. Hence, we choose ZoeDepth for ablation study (\emph{i.e.}, the depth during training and inference is restored with ZoeDepth). Experimental results demonstrate that DINOv2 outperforms ZoeDepth. It is worth noting that ZoeDepth is specifically designed for monocular depth estimation, while DINOv2 is not. We believe that DINOv2 performs better because its extensive and powerful self-supervised pre-training allows it to better extract semantically consistent features. Moreover, we eliminate the diffusion model and retain the same network as the denoiser for feature fusion, followed by using fully connected layers to regress $t$, $R$, and $s$ individually. The significant decline in all metrics demonstrates the effectiveness of the diffusion model.

\begin{table}[t!]
\vspace{-0.5em}
\renewcommand\arraystretch{1}
\newcommand{\tabincell}[2]{\begin{tabular}{@{}#1@{}}#2\end{tabular}}
  \centering
  \caption{Ablation studies for the conditional encoding and denoiser on the REAL275 dataset.}
  \vspace{-0.7em}
  \setlength{\tabcolsep}{4.4pt}{
    \begin{tabular}{c|cc|cc|ccccc}
    \toprule
Row & $G_{f}$ & $T_{s}$ & $T_{b}$ &$S_{c}$ & ${\rm{3}}{{\rm{D}}_{50}}$ & ${\rm{3}}{{\rm{D}}_{75}}$ & \tabincell{c}{$ 10{\rm{cm}}$} & \tabincell{c}{$10^\circ$} & \tabincell{c}{$10^\circ$$ 10{\rm{cm}}$}   \\
    \midrule
    1  & - & \checkmark & \checkmark & \checkmark  & 29.5   &  5.3   & 36.2  & 56.0   & 21.6  \\
    2  & \checkmark & - & \checkmark & \checkmark  & 1.5   &  0.0   & 4.0  & 16.2   & 0.6  \\
    3  & \checkmark & \checkmark & - & -  & 28.6   & 4.8    &  34.9  & 55.5  & 21.7  \\
    4  & \checkmark & \checkmark & - & \checkmark  & 27.7   &  4.5   & 35.9  & 56.2  & 22.6  \\
    5  & \checkmark & \checkmark & \checkmark & -  & 30.9   &  6.0   & 39.5  & 57.1  & 25.0  \\
    6  & \checkmark & \checkmark & \checkmark & *  & 31.3   &  5.2   & 40.1  & \textbf{57.3}  & 25.6  \\
    7  & \checkmark & \checkmark & \checkmark & \checkmark  &  \textbf{31.5}  &  \textbf{6.3}  & \textbf{41.0}  & 56.3  &  \textbf{25.7}  \\
    \bottomrule
    \end{tabular}}
  \label{table4}
  \vspace{-1.5em}
\end{table}

\subsubsection{Conditional Encoding}
We remove the global feature fusion module and the conditional encoding of the time step, respectively, to explore their contribution to MonoDiff9D.

\textbf{Global Feature Fusion ($G_{f}$):}
Compared with the original MonoDiff9D, the 50$\%$ and 75$\%$ $Io{U_{3D}}$ metrics drop by 2.0$\%$ and 1.0$\%$, respectively. The quantitative results are shown in the first row of Table \ref{table4}. These results show that the global feature fusion module in the conditional encoding part enhances the accuracy of MonoDiff9D. Specifically, it helps MonoDiff9D pay more attention to pose-sensitive information from the restored coarse point cloud.

\textbf{Time Step ($T_{s}$):} We remove the conditional encoding of the time step and keep the other encoding part unchanged. The quantitative results are shown in the second row of Table \ref{table4}. The $10{\rm{cm}}$ metric drops from 41.0$\%$ to 4.0$\%$, and the $10^\circ$ metric drops from 56.3$\%$ to 16.2$\%$. These results show that the conditional encoding of the time step is significant for the diffusion model. Since the noise-adding amplitude of the diffusion model correlates with the time step, incorporating time information via conditional encoding can enhance the robustness of the diffusion model learning process.

\subsubsection{Denoiser}
For the diffusion process, we conduct ablation studies on the transformer-based denoiser of the reverse diffusion process to justify its design.

\textbf{Transformer Block ($T_{b}$):}
To prove the applicability of transformer to pose diffusion, we replace the transformer blocks with MLPs and remove the skip concatenation in the denoiser. The $10{\rm{cm}}$ and $10^\circ$ metrics drop by 6.1$\%$ and 0.8$\%$, respectively. When adding the skip concatenation, the $10^\circ$$10{\rm{cm}}$ metric still drops by 3.1$\%$. These results are shown in the third and fourth rows of Table \ref{table4}, which show that the transformer blocks contribute to improving the accuracy of MonoDiff9D. Since the input $I_c$ of the denoiser is a one-dimensional vector, each feature channel in $c_p$ can be globally associated with the condition $c_d$ via the transformer blocks, thus contributing to more precise diffusion.

\begin{table}[t!]
\vspace{-0.5em}
\renewcommand\arraystretch{1}
\newcommand{\tabincell}[2]{\begin{tabular}{@{}#1@{}}#2\end{tabular}}
  \centering
  \caption{Comparing the generalization ability to HS-Pose and IST-Net on the Wild6D \cite{Wild6D} test set.}
  \vspace{-0.7em}
  \setlength{\tabcolsep}{4.6pt}{
      \begin{tabular}{c | c c c c c}
    \toprule
    Method & ${\rm{3}}{{\rm{D}}_{50}}$ & ${\rm{3}}{{\rm{D}}_{75}}$ & \tabincell{c}{$ 10{\rm{cm}}$} & \tabincell{c}{$10^\circ$} & \tabincell{c}{$10^\circ$$ 10{\rm{cm}}$} \\
    \midrule
    %HS-Pose*\cite{HS-Pose}  &  82.1  &  74.7  &  99.6  &  85.4   &  83.7   \\
    DINOv2 + HS-Pose \cite{HS-Pose}  & 8.5   &  0.2  &  33.2  &  11.4  &  3.4   \\
    DINOv2 + IST-Net \cite{IST-Net}  & 3.1   &  0.0  &  26.0  &  15.5  &  7.4   \\
    DINOv2 + MonoDiff9D  &  \textbf{9.6}   &  \textbf{0.4}     &  \textbf{36.3}   &  \textbf{23.4}  &  \textbf{11.1}  \\
    \bottomrule
    \end{tabular}}
  \label{table6}
  \vspace{-1.5em}
\end{table}

\textbf{Skip Concatenation ($S_{c}$):} We first remove it directly to demonstrate the effectiveness of skip concatenation in the denoiser. The quantitative results are shown in the fifth row of Table \ref{table4}. Compared with the original MonoDiff9D, the $10^\circ$$10{\rm{cm}}$ metric drops by 0.7$\%$. Moreover, we use the common residual connection instead of concatenation. The quantitative results are shown in the sixth row of Table \ref{table4}. The the 75$\%$ $Io{U_{3D}}$ metric drops by 1.1$\%$. These results show that skip concatenation in the denoiser is important. Since the skip concatenation can retain more spatial information, it is significant for the reverse diffusion process.

\subsubsection{\textcolor{black}{Generalization in the Wild}}
We use only the test set of the Wild6D \cite{Wild6D} dataset to investigate the generalization ability of MonoDiff9D to images in the wild. Wild6D is a large dataset collected in diverse environments, which includes annotations for 486 test videos featuring various backgrounds and 162 instances. It is worth noting that the Wild6D dataset is mainly used to evaluate self-supervised methods, and its training set has no labels, so we do not train on this dataset. Since current monocular methods either have unavailable codes and models \cite{Synthesis, MSOS, OLD-Net} or require 2.5D sketches as additional input \cite{DMSR}, we use our MonoDiff9D, HS-Pose \cite{HS-Pose}, and IST-Net \cite{IST-Net} to perform ablation studies. Detailed results are shown in Table \ref{table6}. This experiment demonstrates that MonoDiff9D has better generalization ability in the wild than HS-Pose \cite{HS-Pose} and IST-Net \cite{IST-Net}. We believe that this advantage mainly comes from the extensive sampling on the Markov chain of our diffusion-based framework, which can effectively expand the data distribution.

\vspace{-0.5em}
\section{CONCLUSIONS}\label{Conclusion}
\vspace{-0.25em}
This paper presented MonoDiff9D, a diffusion-based method for monocular category-level 9D object pose estimation. We leveraged the probabilistic nature of the diffusion model and the LVM DINOv2 to enhance intra-class unknown object pose estimation when shape priors, CAD models, and depth sensors are unavailable. Extensive experiments demonstrate that MonoDiff9D achieves SOTA accuracy. Future work can use the features of the RGB image to enhance the refinement of the coarse depth map generated by DINOv2. Additionally, exploring the integration of more advanced LVMs could also lead to improved performance in the wild.

\textbf{Acknowledgments:} This work was supported by the National Natural Science Foundation of China under Grant U22A2059 and Grant 62473141, China Scholarship Council under Grant 202306130074, and Natural Science Foundation of Hunan Province under Grant 2024JJ5098.

\bibliographystyle{IEEEtran}
\bibliography{reference}

\end{document}